\documentclass[letterpaper, 10pt, conference]{ieeeconf}
\IEEEoverridecommandlockouts
\usepackage[utf8]{inputenc}
\usepackage{amsmath}
\usepackage{graphicx}
\usepackage{mathptmx}
\usepackage{multirow}
\usepackage{color}
\usepackage{colortbl}
\usepackage{amssymb}
\usepackage[noadjust]{cite}
\usepackage[hyperfootnotes=false]{hyperref}
\usepackage[all]{hypcap}
\usepackage{algpseudocode}
\usepackage{algorithm}
\usepackage{amsmath}
\usepackage{bbold}
\usepackage{tikz}
\usepackage{pgfplots}
\usepackage{pgfplotstable}
\usepackage{comment}
\usepackage{caption}
\usepackage{subcaption}

\usepackage[hyperfootnotes=false]{hyperref}

\def\equationautorefname#1#2\null{%
  (#2\null)%
}

\definecolor{note}{rgb}{0.1,0.1,1}
\definecolor{colgray}{rgb}{0.9275,0.9275,0.9275}
\definecolor{better}{rgb}{.386,0,.627} 
\definecolor{ltbetter}{rgb}{0,.427,.859} 
\definecolor{worse}{rgb}{.572,0,0}

\newcolumntype{g}{>{\columncolor{colgray}}c}

\title{Robot-Supervised Learning for Object Segmentation}
\author{Victoria Florence$^{1}$, Jason J. Corso$^{1,2}$,  and Brent Griffin$^{1,2}$% <-this % stops a space
\thanks{$^{1}$Robotics Institute, University of Michigan}%
\thanks{$^{2}$Electrical Engineering and Computer Science, University of Michigan}%
\thanks{\tt\footnotesize\{vflorenc,jjcorso,griffb\}@umich.edu}%
}

\begin{document}
\bstctlcite{IEEEexample:BSTcontrol}
\maketitle

\begin{abstract}
To be effective in unstructured and changing environments, robots must learn to recognize new objects.
Deep learning has enabled rapid progress for object detection and segmentation in computer vision; however, this progress comes at the price of human annotators labeling many training examples.
% Furthermore, learning-based object detection methods fail to label object classes that are not represented in the training data.
This paper addresses the problem of extending learning-based segmentation methods to robotics applications where annotated training data is not available.
Our method enables pixelwise segmentation of grasped objects.
We factor the problem of segmenting the object from the background into two sub-problems: (1)~segmenting the robot manipulator and object from the background and (2)~segmenting the object from the manipulator.
We propose a kinematics-based foreground segmentation technique to solve (1).
To solve (2), we train a self-recognition network that segments the robot manipulator. We train this network without human supervision, leveraging our foreground segmentation technique from (1) to label a training set of images containing the robot manipulator without a grasped object.
We demonstrate experimentally that our method outperforms state-of-the-art adaptable in-hand object segmentation.
We also show that a training set composed of automatically labelled images of grasped objects improves segmentation performance on a test set of images of the same objects in the environment.
\end{abstract}

\section{Introduction}
%This paper addresses the problem of improving object detection for newly encountered objects using robot-supervised learning.
Although robots are highly productive in controlled environments, developing robotics algorithms that continue to learn new tasks in changing environments is an open problem.
A robust object detector will be indispensable for automation of these tasks, since many industrial and home service tasks require interaction with numerous, ever-changing objects.
Object detection has seen a significant gain in performance in the past decade due to deep learning.
Learning-based methods outperform handcrafted visual features by taking a data-driven approach to generating features that are more robust for object detection~\cite{FasterRCNN,MaskRCNN}. 
However, most deep learning-based methods assume that large quantities of annotated training data are available for each type of object~\cite{MSCOCO,LVIS}, which is impractical when robots encounter new objects and tasks. 
% Furthermore, continuous oversight by a human annotator is cost prohibitive.
Thus, failing to detect new objects is a limitation of fixed-dataset, learning-based detection and a more general obstacle for robot autonomy.

% Autonomous robotics object learning.
For robot perception, simply applying dataset-driven detection methods is wasting a useful asset: robots can take action to improve sensing and understanding of their environments~\cite{IP}.
% Other autonomous robotics object learning.
Various approaches have been created to take advantage of robot embodiment to learn object appearances.
% Some methods rely on specific data-collection environments and are not readily applicable in real-world environments~\cite{zeng,matuszek,vasquez-gomez}. External scene dynamics~\cite{herbst1,herbst2,finman,faulhammer,modayil} as well as pushing objects to create movement~\cite{vanhoof,schiebener,schiebener2} are also used to segment objects from the scene. Other methods perform
We follow the paradigm of past in-hand object segmentation works in which robots grasp unknown objects in order to learn their visual appearance~\cite{browatzki1,browatzki2,krainin1,omrcen,welke, venkataraman}.
%shape or manipulator occlusion.
Most of these methods predate deep learning and require a human to design a visual model or other visual heuristics for recognizing the robot manipulator.
Notably, humans have to redesign these models if there are physical changes to the robot or a new robot is deployed. Work by Browatzki et al. \cite{browatzki1}, which we compare against, and this paper are the only methods we are aware of without this requirement. 

\begin{figure}
\centering
\includegraphics[width=\columnwidth]{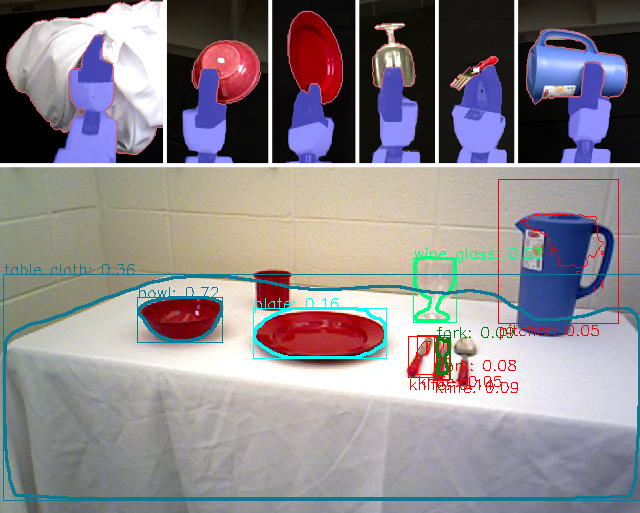}
\caption{Our method produces pixelwise annotations of grasped objects (top). These annotations can be used to improve performance of object instance segmentation methods (bottom). The method adapts to new environments, objects, and robotic platforms without human supervision.} 
\label{fig:frontpage}
\vspace{-1.5em}
\end{figure}

\begin{figure}[p!t]
    \centering
    \includegraphics[width=\columnwidth]{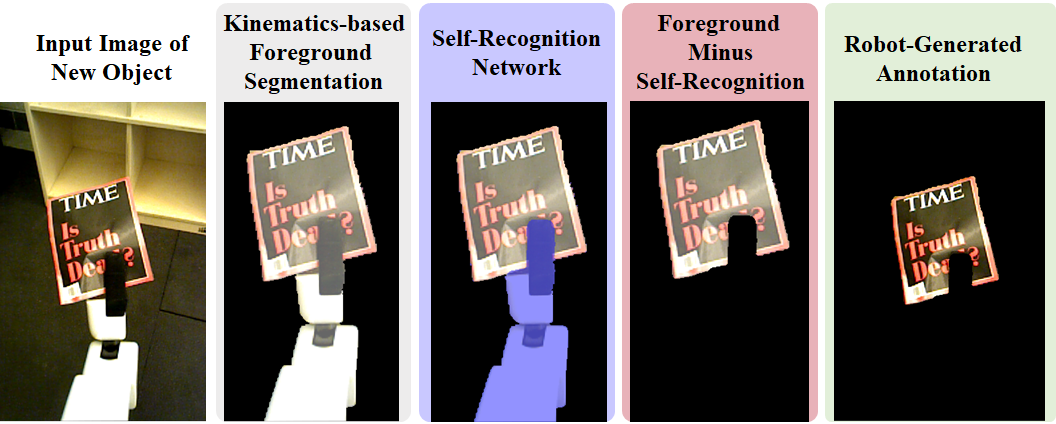}
    \caption{An overview of our method. After collecting images of an object (left), the robot uses encoder readings and active depth sensing to segment the foreground (manipulator or manipulator + grasped object) of each image (middle-left, grey). Using its self-recognition network (middle, blue), the robot isolates the object from the rest of the foreground (middle-right, red). Thus, the robot generates densely-labeled annotations of newly encountered objects (right).}
    \label{fig:pipeline}
    \vspace{-1.5em}
\end{figure}

% Our method overview and combined with overcoming limitations statements.
This paper introduces a method called robot supervision that automatically generates object segmentation training data through robot interaction with grasped objects. In this way, we enable robots to continue improving their own vision systems over time.
% Our method overcomes requirements for handcrafted visual models, object movement, and structured environments by fusing active infrared depth sensing with advances in data-driven models for object detection.
Using only the robot's kinematic link coordinate frames and an RGB-D camera, we segment a grasped object and the manipulator from the background using our kinematics-based foreground segmentation. 
We then separate the robot manipulator from the object using a deep Convolutional Neural Network~(CNN) called a Self-Recognition Network~(SRN), leaving only the in-hand object (see example in Figure~\ref{fig:pipeline}).
Notably, the robot annotates its own training data for the SRN using our kinematics-based foreground segmentation; thus, the SRN can be retrained autonomously.
The end result is a method for generating object labels (like those shown in Figure~\ref{fig:frontpage}) that is generalizable to many existing robot platforms without human supervision. 

% Results
To test our method, we collect and annotate a new dataset that contains RGB-D images of our robot manipulator with 30 in-hand objects (20 images each, 600 total). 
We evaluate our method and that of Browatzki et al.~\cite{browatzki1} on this dataset and show that our method achieves a 27 point (or 75\%) mIoU improvement over the baseline method. 
Finally, we fine-tune an object instance segmentation framework \cite{MaskRCNN} on data produced by our method. The fine-tuning improves object segmentation from 38.1 AP to 49.8 AP on a test set of images. An example of our results and a test image are shown in Figure~\ref{fig:frontpage}. We provide our source code and data at https://github.com/vflorence/RSLOS.
\begin{figure*}[p!t]
\centering
\begin{tikzpicture}[      
        every node/.style={anchor=north west,inner sep=0pt},
        x=1mm, y=1mm,
      ]   
    \node (fighsr) at (0,0)
       {\includegraphics[width=\textwidth]{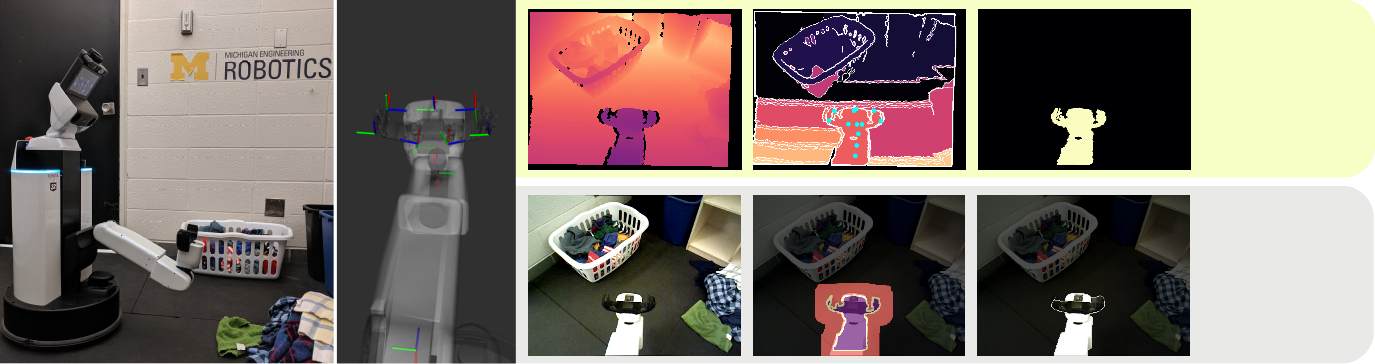}};
    \node (text) [black, text width=1.6cm,align=left,font=\small] at (156,-5){\textbf{Kinematics-based\\ Depth \\Segmentation}};
    \node (text) [black, text width=1.6cm,align=left,font=\small] at (156,-30){\textbf{RGB\\Segmentation\\Refinement}};
\end{tikzpicture}
\caption{In our kinematics-based foreground segmentation method, we use link coordinate frames from the kinematic model of the robot to localize the robot manipulator (left). Using the RGB-D camera's depth channel (top, left), we over-segment the image. We then project manipulator coordinates into the image (top, middle) and label the segments containing filtered projected coordinates as foreground (top, right). 
We refine the segmentation with the RGB channels (bottom, left), initializing GrabCut~\cite{rother} with a depth segmentation-based mask (bottom, middle) to give us our final output (bottom, right).
}
\label{fig:KIFS}
\vspace{-1.5em}
\end{figure*}

\section{Related Work}

\subsection{Interactive Object Segmentation}
In interactive object segmentation methods, robot actions or environment continuity are used in order to segment the object from the rest of the image.
Some methods learn objects that are placed in uncluttered or known environments~\cite{zeng,matuszek,vasquez-gomez}. Other works rely on scene change over time, calling anything that violates the static scene assumption an object~\cite{herbst1,herbst2,finman,faulhammer,modayil}. They rely on the non-guaranteed movement of objects. Other methods in this category push objects to create movement in the scene and group pixels that move together into object labels~\cite{vanhoof,schiebener,schiebener2}. These methods have more control over object movement, but they require a work surface and objects that permit pushing. 
The benefit of these methods is that they can segment many objects at the same time; however, they do not allow for full pose or environmental control. 

\subsection{In-hand Object Segmentation} 
In-hand object segmentation methods use robot encoder feedback to locate a grasped object and various methods to reason about robot-object occlusion. Omr\v{c}en et al.~\cite{omrcen} incorporate many data sources such as color distribution, a disparity map, and a pretrained Gaussian Mixture Model (GMM) of the hand to segment unknown objects, but their method does not extract pixelwise object labels. Welk et al.~\cite{welke} use Eigen-backgrounds, disparity mapping, and tracking on a model of the robot manipulator to isolate objects. Krainin et al.~\cite{krainin1} take a self-recognition based approach to object learning by matching a robot manipulator to its 3D mesh model in order to isolate and model an in-hand object. However, their method requires a 3D geometric model of the robot manipulator and focuses on modeling non-deformable objects from multiple viewpoints.  Browatzki et al.~\cite{browatzki1,browatzki2} use a GMM trained on pixel values around a bounding box to isolate a grasped object. This method focuses on viewpoint selection and data association, but their object isolation technique is not robust to pixel-level similarity between the object, background, and robot. While these systems are able to isolate unseen objects for visual learning and oftentimes model the occlusion of the robot manipulator by these objects, they are limited by the need for custom manipulator models, environment-specific heuristics, and parameter tuning.
 
\subsection{Self-supervised Manipulator Recognition} 
To the best of our knowledge, da Costa Rocha et al.~\cite{dacostarocha} is the first method to ``make use of the kinematic model of a robot in order to generate training labels." This work learns pixelwise self-recognition of the da Vinci surgical robot with unsupervised training labels much like our SRN. da Costa Rocha's method uses a projection of a full geometric model of the manipulator into an RGB image to generate the labels, while our method requires depth sensor readings and only uses link coordinate frames (e.g., in Figure~\ref{fig:KIFS}). The benefit of our approach is that we learn self-recognition without a full model of the robot and adapt to changes in robot hardware automatically. Additionally, the end goal of our method is in-hand object isolation for object learning, while da Costa Rocha's work focuses on robotic self-recognition.

\section{Kinematics-based Foreground Segmentation}
\label{sec:fgseg}
\subsection{Kinematics-based Depth Segmentation}
We begin segmenting the robot's manipulator from background by over-segmenting the head camera's depth image. We use $D \in \mathbb{R}^{H \times W}$ to represent the depth image, where $H$ and $W$ are the numbers of rows and columns in $D$. $D(i,j) \in \mathbb{R}$ is the measured depth in millimeters for the pixel at row $i$, column $j$. Using the graph-based image segmentation of Felzenswalb~\cite{felzenszwalb}, we define the depth-image segmentation 
\begin{align}
    S(D) := \{ s_1, s_2, \ldots, s_n \},
    \label{eq:dseg}
\end{align}
where $S$ is exhaustive of the 2D coordinates in $D$ with mutually exclusive subsets.

Next, we use the robot's encoder readings and kinematic model to get approximate 3D coordinates of link positions and project them into the depth image.
We take the 3D link position of each $i$th link relative to the camera's coordinate frame, $\text{P}_i := [\text{x}_i, \text{y}_i, \text{z}_i]^\top$, and find the projected depth-image coordinates using the transform
\begin{align}
    \label{eq:jointproj}
    [p_i, 1]^\top := \lfloor  \frac{\mathbf{K} \text{P}_i}{\text{z}_{i}} + 0.5 \rfloor = [h_i, w_i, 1]^\top
    ,
\end{align}
where $\mathbf{K}$ $\in \mathbb{R}^{3\times3}$ is the head camera's intrinsic camera matrix.

%\begin{align}
%    \forall j_i, p_i:=p_{(j_{i1}, j_{i2})} \Longleftarrow p_i \in D.
%\end{align}

Using the depth segmentation~\eqref{eq:dseg} and projected kinematic points~\eqref{eq:jointproj}, we define our initial foreground segmentation
\begin{align}
\label{eq:depmask}
s_\text{fg} := \bigcup \{ s_j | \exists p_i \in s_j, D(h_i,w_i) \leq \text{z}_i + \lambda\},
\end{align}
where $D(h_i,w_i)$ is the depth measurement corresponding to $p_i$ and $\lambda$ is a noise threshold for $\textrm{z}_i$.
Put simply, if the depth sensor reading ($D(h_i,w_i)$) of a projected link location ($p_i$) is within $\lambda$ (distance) past its expected kinematic location ($\text{z}_i$), the segment ($s_j$) containing that reading is added to the initial foreground segmentation ($s_\text{fg}$).

\subsection{RGB Segmentation Refinement}
% \subsection{Morphology- and RGB-based Segmentation Refinement}
%\subsection{Segmentation Refinement via Matrix Operations and RGB}
We refine the initial foreground segmentation using matrix operations. 
We represent $s_\text{fg}$ \eqref{eq:depmask} as matrix $\mathbf{M}_0 \in \mathbb{R}^{H \times W}$ s.t.
%A matrix representation of $C_\text{fg}$ called $\mathbf{M}_\text{orig} \in \mathbb{R}^{h \times w}$ is defined as
\begin{align}
    \label{eq:origmask}
    \mathbf{M}_{0}(i,j) :=
        \begin{cases}
            1 & D(i,j) \in s_\text{fg}\\
            0 & \text{otherwise}
        \end{cases},
\end{align}
where $\mathbf{M}_{0}(i,j)=1$ is a foreground element. We post-process $\mathbf{M}_0$ by filling in all holes then performing a morphological binary opening operation with the kernel $\textbf{J}_8$ where $\textbf{J}_N \in \mathbb{R}^{N \times N}$ and every element is equal to one.
These operations reduce the effect of depth sensor noise, which often manifests as holes and noisy object edges.

Using the processed $\mathbf{M}_0$, we generate two more matrices
\begin{align}
    \label{eq:errmask}
    \mathbf{M}_\text{p} := &~ \text{erosion}(\mathbf{M}_0,\textbf{J}_{10}),\\
    \label{eq:dilmask}
    \mathbf{M}_\text{r} := &~ \text{dilation}(\mathbf{M}_0,\textbf{J}_{75}),
\end{align}
where $\mathbf{M}_\text{p}$ is precision oriented, and $\mathbf{M}_\text{r}$ is recall oriented. Note that $\mathbf{M}_{\text{p}}=1 \implies \mathbf{M}_{\text{0}}=1$, and $\mathbf{M}_{\text{0}}=1 \implies \mathbf{M}_{\text{r}}=1$.
The erosion and dilation operations soften the boundary created by the depth segmentation to account for depth sensor noise and any misalignment between the RGB and depth images.

We generate our final foreground segmentation using a GrabCut segmentation~\cite{rother} on the RGB image $I \in \mathbb{R}^{H \times W}$ corresponding to $D$.
Using $\mathbf{M}_\text{p}$, $\mathbf{M}_\text{r}$, and the processed $\mathbf{M}_0$, we initialize the GrabCut segmentation using
\begin{align}
    \label{eq:gcmask}
    \mathbf{M}_{\text{gc}} := \mathbf{M}_{\text{r}} + \mathbf{M}_{0} + \mathbf{M}_{\text{p}},
\end{align}
where $\mathbf{M}_{\text{gc}}(i,j) \in \{0,1,2,3\}$, 0 is background, 1 is probably background, 2 is probably foreground, and 3 is foreground.
We refine $\mathbf{M}_\text{gc}$ using GrabCut for 8 iterations and convert the refined $\mathbf{M}_\text{gc}$ to the binary mask
\begin{align}
    \label{eq:binary}
    \mathbf{M}_{\text{fg}}(i,j) :=
        \begin{cases}
            0 & \mathbf{M}_{\text{gc}}(i,j) \in \{0,1\} \\
            1 & \mathbf{M}_{\text{gc}}(i,j) \in \{2,3\}
        \end{cases}.
\end{align}
$\mathbf{M}_\text{fg}$ is the final kinematics-based foreground segmentation mask, where $\mathbf{M}_{\text{fg}}(i,j)=1$ indicates that $I(i,j)$ corresponds to the robot's manipulator in the foreground (see example in Figure~\ref{fig:KIFS}).

\section{Robot-supervised Self-recognition Network}
\label{sec:selfnet}
Using the kinematics-based foreground segmentation described in Section~\ref{sec:fgseg}, we enable the robot to collect and label its own data to train an SRN that labels instances of the robot manipulator in an image.
Specifically, the robot performs foreground segmentation on images where its manipulator is the only object in the foreground (i.e., no objects are grasped) and creates a manipulator annotation from the foreground mask.

To diversify training data for the SRN, we collect images of the robot manipulator in various poses, permuting across all combinations of individual joint positions.
Poses are uniformly distributed across each joint's range, and the number of positions per joint can be set as a parameter to match the time available for learning. 
For each pose we 1) position the camera such that the robot gripper is approximately centered in the image and 2) command the non-varied manipulator joints to a position that puts the gripper beyond the depth camera's minimum range.
The data collection joint configuration can be adjusted to fit any robot platform, as long as these two requirements are met.

\begin{figure}
\centering
\includegraphics[width=.9\columnwidth]{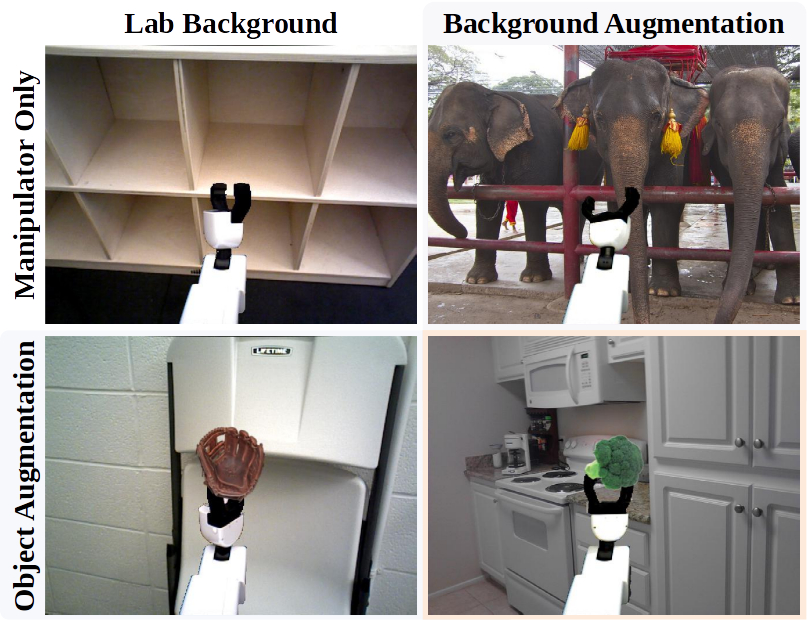}
\caption{In the self-recognition network training set, we augment manipulator images with background images taken by the robot (left) and from the COCO dataset (right) as well as foreground objects (bottom). These augmentations increase the diversity of the background and manipulator classes, respectively, and increase the number of example images in the self-recognition network training set.
}
\label{fig:backgroundsub}
\vspace{-1.5em}
\end{figure}

We use $\mathbf{M}_\text{fg}$ \eqref{eq:binary} to label each manipulator image as a training example, where the background class ID is 0 and the manipulator class ID is 1. 

Inspired by work on domain randomization for sim-to-real transfer learning~\cite{tobin},~\cite{tremblay}, we perform dataset augmentation to close the gap between the automatically labeled manipulator-only images and future test-time images with objects in-hand. We did this with background substitution and foreground object augmentation. Examples of these augmentations can be seen in Figure~\ref{fig:backgroundsub}.

We do background substitution with background images taken from the popular Common Objects in Context (COCO) dataset~\cite{MSCOCO} as well as pictures taken by the robot without the manipulator in view.

We perform foreground superposition with object classes in the COCO dataset~\cite{MSCOCO}. For each foreground augmentation, we randomly scale and rotate an alpha-matted object image and overlay it at the center of the image.

The background and foreground augmentations can increase the number and type of objects sampled in the training background class, while the foreground augmentations randomly alter manipulator appearance to imitate robot-object occlusion. Increasing dataset diversity and size reduces overfitting and improves the robustness of the learned manipulator segmentation. 

We use an object instance segmentation framework, Mask R-CNN, pretrained on the COCO dataset with R-50-FPN backbone as the SRN model. We use the standard multi-task loss from the original paper \cite{MaskRCNN}. 

\begin{figure}
\centering
\includegraphics[width=\columnwidth]{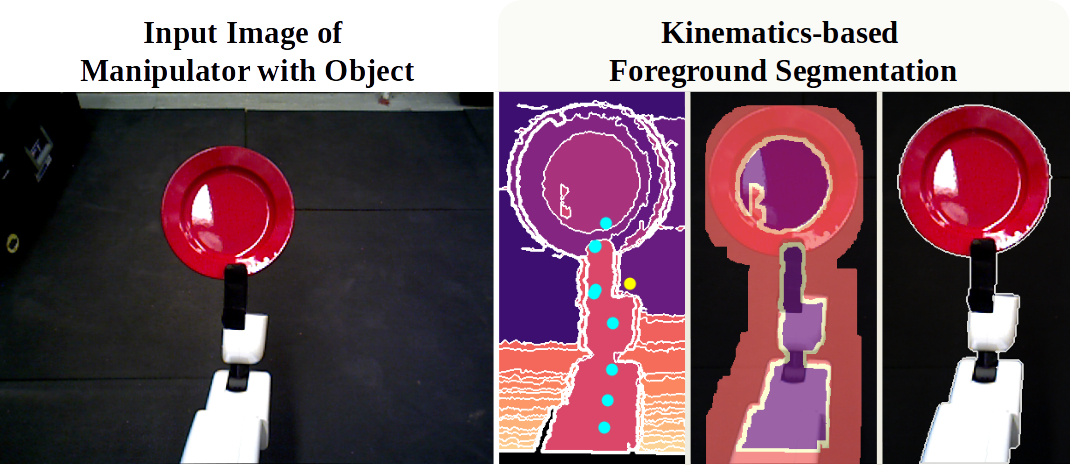}
\caption{We use kinematics-based foreground segmentation to segment the gripper and object from the background. The output (right) is combined with the SRN output to create robot-supervised object annotations.}
\label{fig:obj_seg}
\vspace{-1.5em}
\end{figure}

\section{Robot-supervised Object Annotation}
%\section{Robot-generated Annotations of Objects}
%\subsection{Object segmentation}
\label{sec:objant}

In order to annotate in-hand objects, we repeat the data collection procedure from Section~\ref{sec:selfnet} with objects grasped by the robot manipulator. We apply our kinematics-based foreground isolation method to create $\mathbf{M}_\text{fg}$ \eqref{eq:binary} for each image (see example in Figure~\ref{fig:obj_seg}). The SRN is applied to predict the manipulator location. In cases where the SRN returns multiple predictions, we take the prediction with the highest score. We call the mask representing the SRN prediction $\mathbf{M}_\text{SRN}$.
To create the final object labels, $\mathbf{M}_\text{fg}$ and $\mathbf{M}_\text{SRN}$ are combined for each image as
\begin{align}
    \label{eq:objmask}
    \mathbf{M}_\text{object} = \mathbf{M}_\text{fg} \bigcap \lnot \mathbf{M}_\text{SRN}.
\end{align}

We then perform a final opening on $\mathbf{M}_\text{object}$ with $\textbf{J}_3$, a matrix of ones, for noise removal. 
Examples of this process can be seen in Figure~\ref{fig:qualresults}, where $\mathbf{M}_\text{SRN}$ is shaded blue and $\mathbf{M}_\text{object}$~\ref{eq:objmask} is outlined in red.

\section{Experiments}
\subsection{Experimental Setup}
All experiments use the Toyota Human Support Robot (HSR) \cite{HSR_journal}. Images are gathered with its RGB-D head camera. In the object learning experiments, the robot begins with objects grasped; recent works such as~\cite{pinto} and~\cite{mahler} demonstrate viable methods for unknown object grasping. 
We use a Pytorch implementation of Mask R-CNN for the SRN and object re-recognition networks~\cite{massa,pytorch}. 

\subsection{Metrics}
On the grasped object annotation task, we use pixelwise mIoU as our metric. For each image, we create a ground-truth object annotation that is compared to our method's output as well as the baseline. We calculate a standard, per-image, Intersection over Union (IoU) metric for the object masks and average the results for each class to get a class mean IoU (mIoU). The overall mIoU is an average of all class mIoUs. 

On the object re-recognition in context task, we use the COCO API Average Precision (AP) detection metrics for evaluating performance on this task~\cite{MSCOCO}. AP is calculated as the area under the precision recall curve. We provide results for different IoU and pixels-per-object thresholds. For more details about these metrics, we recommend looking at the COCO detection task evaluation metric \cite{MSCOCO}.

The difference in metric is motivated by the fact that the output for the first task is a single mask with labels for each pixel, while the output for the second task can include multiple detections per image region.

\begin{figure}[t]
\centering
\includegraphics[width=0.9\columnwidth]{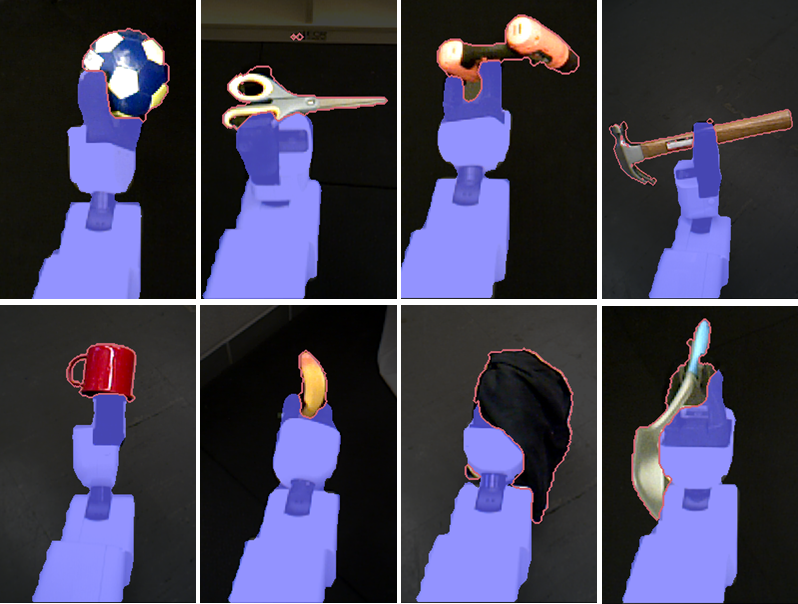}
\caption{Some of our method's best qualitative results. The areas shaded blue are labeled as manipulator by the SRN, while the red-outlined areas are the object labels. These pixelwise segmentations of in-hand objects can be used to train data-driven object detectors.}
\label{fig:qualresults}
\end{figure}
\subsection{Grasped Object Annotation Performance}
To quantify the performance of our robot-supervised object annotation method, we gather a test set of 600 human-annotated images of the manipulator with an object grasped. 
For each of 30 objects from the Yale-CMU-Berkeley (YCB) standardized object dataset~\cite{YCB} we collect 20 images of the grasped object. The object viewing poses are uniformly sampled from the joint spaces of two revolute joints in what could be called the ``wrist" of the robot; this configuration allows for coverage of the entire viewing sphere of the grasped object without consideration of robot-object occlusion. Due to robot-object occlusion, it is not possible to view the entire object from a single grasp, and the object is not guaranteed to be visible in every image. 

For our method, we use Felzenswalb segmentation parameters of $\sigma$~=~0.5 and k~=~800, which are approximately scaled by image size from the parameter settings in the original paper~\cite{felzenszwalb}. Additionally, we set the depth segmentation noise threshold to $\lambda=200~\textrm{mm}$. The SRN used in our experiments is trained and validated on 208 and 89 images of HSR's manipulator, respectively. We annotate all images automatically with the kinematics-based foreground segmentation method (Section~\ref{sec:fgseg}) and augment them with the methods described in Section~\ref{sec:selfnet}. The number of background and foreground images that we use is shown in Table~\ref{table:collecteddata}. For all SRN training, we follow the Detectron solver scheduling rules~\cite{detectron}.

\subsubsection{Ablation of Data Augmentation}
To show the contribution of each type of augmentation in our SRN dataset creation method, we train multiple SRNs, each using a different combination of data augmentations. We repeat each type of data augmentation three times per the most plentiful augmentation resource involved, repeating resources with fewer elements as necessary.
For example, we have more backgrounds than gripper images, so to create the ``BG" dataset, we use each background 3 times and repeat the gripper images as necessary. The number of images in each augmented data split is shown in Table~\ref{table:dataset}. 

Results from the ablative SRNs show that foreground (FG) and background (BG) augmentations add 2.9 and 4.7 point improvements, respectively, to our SRN performance as shown in \autoref{table:ablation}. The combined foreground, background augmentation add a 6.5 point improvement. Results show that the object recall scores consistently benefit from augmentations. Improved recall scores indicate that there are fewer false negatives on the object labeling task (i.e. fewer false positives by the SRN).

By adding data examples through our data augmentation methods, we are able to increase the precision of the SRN detector and achieve higher recall on object segmentation.

\setlength{\tabcolsep}{6pt}
\begin{table}[t]
	\centering
	\caption{Data Collection}
	\begin{tabular}{| r | c | c | c |}
		\hline 
		\multicolumn{1}{|c|}{Data}	&	Total	&	Train	&	Val	\\
		\hline
        Gripper Images	&	297	&	208	&	89	\\
        \hline
        Background Images	&	1800	&	1260	&	540	\\
        (COCO/Lab)	&	(1746 / 54)	&	(1220 / 40)	&	(526 / 14)	\\
        \hline
        Foreground Objects	&	80	&	56	&	24	\\
		\hline
	\end{tabular}
	\label{table:collecteddata}
\end{table}

\setlength{\tabcolsep}{6pt}
\begin{table}[t]
	\centering
	\caption{Augmented Datasets}
	\begin{tabular}{| r | c |}
		\hline 
        \multicolumn{1}{|c|}{Dataset}	&	Total	\\
        \hline
        \textbf{Orig} - original images &	297	\\
        \textbf{FG} - foreground augmentation &	891	\\
        \textbf{BG} - background augmentation &	5400	\\
        \textbf{FGBG} - foreground, background augmentation	&	5400	\\
        \textbf{Ours} - all data &	11988	\\
	    \hline
	\end{tabular}
	\label{table:dataset}
	\vspace{-1.5em}
\end{table}

\subsubsection{Robot-Supervised Object Annotation Performance}

We compare our method to the prior work of Browatzki et al.~\cite{browatzki1}. Their work focuses on efficient viewpoint selection for object learning and included an in-hand object segmentation method. Similarly to our method, it does not require human-designed vision heuristics for reasoning about robot-object occlusion. The method trains a Gaussian Mixture Model on the pixels within a frame around the expected object location (i.e. between inner and outer bounding boxes). We re-implement their segmentation method and choose parameters based on a parameter sweep over the test set. (bounding box sizes~=~(270,300), number of Gaussians~=~1, threshold~=~1E-15).

The overall mIoU results in \autoref{table:ablation} indicate that our method outperforms the baseline on the task of object segmentation. Additionally, our method outperforms \cite{browatzki1} on the majority of objects as shown in Figure~\ref{table:objectmIoU}. Examples of segmentation results for our method are shown in \autoref{fig:qualresults}. In order to gain insight on the performance of each method, we look at the performance along the axes of pixels per object and average saturation. Data regarding the relationship between object size and method performance can be viewed in Figure~\ref{fig:miouvpx}, while method performance versus object saturation is shown in Figure~\ref{fig:miouvsat}. Note that our robot platform, HSR, is black and white, so saturation is a pixel-based metric representing robot-object visual similarity. 

Overall, this experiment indicates increased robustness of our method over the prior work on in-hand object segmentation without human-designed, robot-specific heuristics for reasoning about robot-object occlusion. Our method enables us to apply advances in deep learning based object-segmentation without human annotation and demonstrates improved accuracy over the pixel-based method in Browatzki et al.~\cite{browatzki1}.
\setlength{\tabcolsep}{3pt}
\begin{table}[t]
    \centering
    \caption{Method Comparison and Ablation Study}
    \begin{tabular}{|l|c |c|c|}
        \hline
        \multicolumn{1}{|c|}{Method} 	&		mIoU		&		Precision		&		Recall		\\
        \hline
Orig	&		0.431		&		0.788		&		0.477		\\
FG	&		0.460		&		0.780		&		0.516		\\
BG	&		0.478		&		0.784		&		0.531		\\
FGBG	&		0.506		&		0.795		&		0.563		\\
Ours	&	\textbf{	0.639	}	&	\textbf{	0.823	}	&		\textbf{0.716}		\\
Browatzki et al. \cite{browatzki1}	&		0.362		&		0.685		&		0.400		\\

        \hline
    \end{tabular}
    \label{table:ablation}
\end{table}

\setlength{\tabcolsep}{4pt}
\begin{table}
	\centering
	\caption{Object Segmentation Performance (mIoU)}
	\def\arraystretch{1.05}
	\begin{tabular}{| c | c | c |}
	\hline	
	Object	&		Browatzki et al.~\cite{browatzki1}	& Ours	\\
\hline \hline										
\rowcolor{better!2}	Pitcher	&		0.934		&	\textbf{	0.957	}	\\
\rowcolor{better!9}	Bowl	&		0.864		&	\textbf{	0.951	}	\\
\rowcolor{better!6}	Mug	&		0.867		&	\textbf{	0.931	}	\\
\rowcolor{better!72}	Wood	&		0.184		&	\textbf{	0.909	}	\\
\rowcolor{better!52}	Magazine	&		0.363		&	\textbf{	0.885	}	\\
\rowcolor{worse!2}	Apple	&	\textbf{	0.895	}	&		0.875		\\
\rowcolor{better!15}	Brick	&		0.716		&	\textbf{	0.865	}	\\
\rowcolor{better!42}	Plate	&		0.445		&	\textbf{	0.860	}	\\
\rowcolor{better!5}	Soccer ball	&		0.802		&	\textbf{	0.852	}	\\
\rowcolor{better!22}	Power drill	&		0.625		&	\textbf{	0.844	}	\\
\rowcolor{better!83}	Tshirt	&		0.002		&	\textbf{	0.831	}	\\
\rowcolor{better!76}	Wine glass	&		0.054		&	\textbf{	0.814	}	\\
\rowcolor{better!46}	Scissors	&		0.260		&	\textbf{	0.721	}	\\
\rowcolor{better!28}	Hammer	&		0.428		&	\textbf{	0.708	}	\\
\rowcolor{better!46}	Screwdriver	&		0.240		&	\textbf{	0.696	}	\\
\rowcolor{better!57}	Rope	&		0.106		&	\textbf{	0.676	}	\\
\rowcolor{worse!18}	Banana	&	\textbf{	0.837	}	&		0.654		\\
\rowcolor{better!37}	Padlock	&		0.177		&	\textbf{	0.546	}	\\
\rowcolor{better!13}	Fork	&		0.407		&	\textbf{	0.537	}	\\
\rowcolor{better!8}	Spoon	&		0.411		&	\textbf{	0.493	}	\\
\rowcolor{better!40}	Wrench	&		0.092		&	\textbf{	0.489	}	\\
\rowcolor{better!38}	Expo marker	&		0.107		&	\textbf{	0.486	}	\\
\rowcolor{better!32}	Spatula	&		0.090		&	\textbf{	0.413	}	\\
\rowcolor{better!41}	Tablecloth	&		0.003		&	\textbf{	0.408	}	\\
\rowcolor{better!3}	Knife	&		0.374		&	\textbf{	0.406	}	\\
\rowcolor{better!24}	Keys	&		0.106		&	\textbf{	0.345	}	\\
\rowcolor{better!22}	Baseball	&		0.071		&	\textbf{	0.291	}	\\
\rowcolor{better!21}	Golf ball	&		0.079		&	\textbf{	0.289	}	\\
\rowcolor{better!12}	Dice	&		0.101		&	\textbf{	0.223	}	\\
\rowcolor{worse!2}	Sponge	&	\textbf{	0.228	}	&		0.210		\\
										
\hline
	Averaged	&		0.362		&	\textbf{	0.639	}	\\
        \hline
	\end{tabular}
	\label{table:objectmIoU}
	\vspace{-1.5em}
\end{table}

\pgfplotsset{
    mark1/.style={ltbetter,only marks,mark=*},
    line1/.style={ltbetter,thick},
    mark2/.style={worse,only marks,mark=square*},
    line2/.style={worse,thick},
    combo legend 1/.style={
        legend image code/.code={
            \draw [/pgfplots/mark1] plot coordinates {(1mm, 0cm)};
            \draw plot coordinates {(2.5mm, -3pt)} node {,};
            \draw [/pgfplots/line1] plot coordinates {(4.3mm, 0mm)(7.3mm, 0mm)};
        },
        font=\Large
    },
    combo legend 2/.style={
        legend image code/.code={
            \draw [/pgfplots/mark2] plot coordinates {(1mm, 0cm)};
            \draw plot coordinates {(2.5mm, -3pt)} node {,};
            \draw [/pgfplots/line2] plot coordinates {(4.3mm, 0mm)(7.3mm, 0mm)};
        },
    },
    label style={font=\Large},
}
\pgfplotstableread{ObjMiou.txt}\objmiou

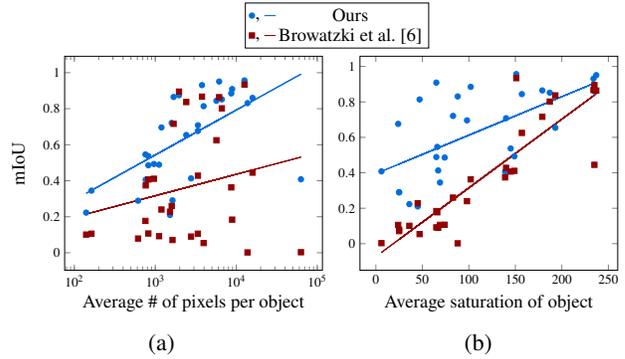
\begin{figure}
    \centering
\begin{subfigure}[b]{0.23\textwidth}
    \begin{tikzpicture}[scale=0.5]
    \begin{semilogxaxis}[
        legend style={at={(0.7,1.25)},anchor=north west,font=\Large},
        xlabel=Average \# of pixels per object, 
        ylabel=mIoU
        ]
        
        \addplot[forget plot, mark1, only marks] table[x=px, y=our]{\objmiou};
        \addplot[forget plot, line1] table[x=px, y={create col/linear regression={y=our}}]{\objmiou};
        \addlegendimage{combo legend 1};
        
        \addplot[forget plot, mark2, only marks] table[x=px, y=bro]{\objmiou};
        \addplot[forget plot, line2] table[x=px, y={create col/linear regression={y=bro}}]{\objmiou};
        \addlegendimage{combo legend 2};

        \legend{Ours,Browatzki et al. \cite{browatzki1}};
    \end{semilogxaxis}
    \end{tikzpicture}
    \caption{}
    \label{fig:miouvpx}
\end{subfigure}
\begin{subfigure}[b]{0.23\textwidth}
    \centering
    \begin{tikzpicture}[scale=0.5]
    \begin{axis}[
        legend pos=north west,
        xlabel=Average saturation of object
        ]
        \addplot [forget plot, mark1, only marks] table[x=sat, y=our]{\objmiou};
        \addplot [forget plot, line1] table[x=sat, y={create col/linear regression={y=our}}]{\objmiou};
        \addlegendimage{combo legend 1};
        
        \addplot [forget plot, mark2, only marks] table[x=sat, y=bro]{\objmiou};
        \addplot [forget plot, line2] table[x=sat, y={create col/linear regression={y=bro}}]{\objmiou};
        \addlegendimage{combo legend 2};
    \end{axis}
    \end{tikzpicture}
    \caption{}
    \label{fig:miouvsat}
\end{subfigure}
\caption{We examine our method's performance along two pixel-based metrics of image content. On the axis of pixels per object mask, we observe that the baseline method fails on very large objects, and that the performances of both methods decrease on small objects (left). Additionally, we observe that our method's performance was slightly less correlated with  pixel-level similarity to the manipulator (right).}
\label{fig:mioupxsat}
\end{figure}

\setlength{\tabcolsep}{3pt}
\begin{table}[b]
    \vspace{-1.5em}
    \centering
    \caption{Object Instance Segmentation Performance}
    \begin{tabular}{|c|c|c|c|c|c|c|}
        \hline
        Dataset	&	$\text{AP}$	&	$\text{AP}^{0.50}$	&	$\text{AP}^{0.75}$	&	APs 	&	APm 	&	APl 	\\
        \hline
        \multicolumn{7}{|c|}{Bounding Box} \\ \hline

        LVIS	&	39.2	&	51.0	&	45.6	&	35.5	&	44.3	&	62.3	\\
        LVIS + fine-tuning on our data	&	49.9	&	70.8	&	57.6	&	50.6	&	57.3	&	63.3	\\									\hline
        \multicolumn{7}{|c|}{Segmentation}\\
        \hline
        LVIS	&	38.1	&	50.9	&	45.1	&	30.8	&	42.4	&	61.3	\\
        LVIS + fine-tuning on our data	&	49.8	&	69.7	&	54.7	&	36.6	&	57.0	&	65.4	\\
        \hline
    \end{tabular}
    \label{table:rerec}
\end{table}

\subsection{Object Re-recognition in Context}
In order to analyze the usefulness of the labels produced by our method, we compare a state-of-the-art object instance segmentation framework trained on dataset images to one that is fine-tuned on outputs from our method. The same model, initialization, and training schedules are used as in the SRN training procedure for both networks.
As a baseline, we train Mask R-CNN on a subset of the Large Vocabulary Instance Segmentation dataset v0.5 (LVIS v0.5) by Gupta et al.~\cite{LVIS} that has 25 classes in common with our selected YCB objects (excluding brick, dice, golfball, rope, and wood). We reduce the LVIS v0.5 dataset to the images containing any of the 25 YCB objects and remove all other classes from the annotations, maintaining the training and validation splits. 
For our method, we collect and automatically label 20 additional images per object for the 25 overlapping classes. We then fine-tune the LVIS-trained network on our method's output labels, leaving out approximately 8\% of the object images as a validation set. Results in Table~\ref{table:rerec} show that our method outperforms the dataset baseline on all metrics. 
This result indicates that despite the noise we observe in the in-hand object segmentation results in Figure~\ref{table:objectmIoU}, the limited training examples of in-hand objects are useful for training a deep CNN to recognize the same objects in different contexts. 

\section{Conclusions}
We have presented and experimentally validated robot supervision that enables robots to generate new annotations using in-hand object segmentation. Our method of kinematics-based foreground segmentation followed by a robot-supervised SRN achieves significant improvement over the baseline on the task of in-hand object segmentation. Our method performs well on a wide variety of objects and is not specific to a single robot platform. Additionally, experiments indicate that fine-tuning an object instance segmentation framework on labels created by our method improves performance on object segmentation in context. Using our method, robots can generate their own training data and learn to better segment new objects and environments without human supervision. 

\section*{ACKNOWLEDGMENT}
Toyota Research Institute (``TRI'') provided funds to assist the authors with their research but this article solely reflects the opinions and conclusions of its authors and not TRI or any other Toyota entity.

\bibliographystyle{IEEEtran} 
\bibliography{IEEEabrv,refs}
\end{document}